\documentclass[letterpaper, 10 pt, conference]{ieeeconf}
\usepackage{ifpdf}
\usepackage{amsmath,amsfonts}
\usepackage{amssymb}  
\usepackage{algorithmic}
\usepackage{algorithm}
\usepackage{array}
\usepackage{multirow}

\usepackage[caption=false,font=footnotesize]{subfig}
\usepackage{textcomp}
\usepackage{stfloats}
\usepackage{url}
\usepackage{verbatim}
\usepackage{soul}
\usepackage{graphicx}
\usepackage{cite}
\usepackage{booktabs}
\usepackage[export]{adjustbox}
\usepackage{bm}
\usepackage{nccmath}
\usepackage{mathtools}
\usepackage{tensor}
\usepackage[dvipsnames]{xcolor}
\usepackage{units}
\usepackage{tikz-cd}
\usepackage{bbm}
\usepackage{tikz}
\usepackage{cases}
\usepackage{epsfig}

\usepackage{hyperref}

\begin{document}

\title{Whole-body Bilateral Teleoperation for Dynamic Mobile Lifting Tasks of a Wheeled Humanoid}
\title{Dynamic Lifting Tasks via Bilateral Teleoperation of a Wheeled Humanoid}
\title{Teleoperated heavy lifting tasks of a wheeled humanoid}
\title{Heavy lifting tasks via haptic teleoperation of a wheeled humanoid}

\author{{Amartya Purushottam$^1$, Jack Yan$^{2}$, Christopher Xu$^1$, and Joao Ramos$^{1,2}$}
\thanks{$^{1,2}$The authors are with the $^1$Department of Electrical and Computer Engineering and the $^2$Department of Mechanical Science and Engineering at the University of Illinois at Urbana-Champaign, USA.}
\thanks{This work is supported by the National Science Foundation via grant IIS-2024775.}}

\maketitle

\begin{abstract}

Humanoid robots can support human workers in physically demanding environments by performing tasks that require whole-body coordination, such as lifting and transporting heavy objects. These tasks, which we refer to as Dynamic Mobile Manipulation (DMM), require the simultaneous control of locomotion, manipulation, and posture under dynamic interaction forces. This paper presents a teleoperation framework for DMM on a height-adjustable wheeled humanoid robot for carrying heavy payloads. A Human-Machine Interface (HMI) enables whole-body motion retargeting from the human pilot to the robot by capturing the motion of the human and applying haptic feedback. The pilot uses body motion to regulate robot posture and locomotion, while arm movements guide manipulation. Real-time haptic feedback delivers end-effector wrenches and balance-related cues, closing the loop between human perception and robot-environment interaction. We evaluate different telelocomotion mappings that offer varying levels of balance assistance, allowing the pilot to either manually or automatically regulate the robot’s lean in response to payload-induced disturbances. The system is validated in experiments involving dynamic lifting of barbells and boxes up to 2.5 kg (21\% of robot mass), demonstrating coordinated whole-body control, height variation, and disturbance handling under pilot guidance. 
\\Video demo can be found at: \\
\url{https://youtu.be/jF270_bG1h8?feature=shared}
\end{abstract}


\IEEEpeerreviewmaketitle

\section{Introduction}

Humanoid robots show promise in accomplishing physically demanding tasks in human-centered environments, such as warehouses and grocery stores. To be practically deployable, they must perform human-like behaviors reliably. For example, a warehouse worker may need to bend down to pick up a heavy object from the ground and then lift and place it onto a shelf - an action requiring simulatenous locomotion, and manipulation. Executing such tasks require humanoid robots to coordinate their whole-body motion while regulating interaction forces, and contact states — particularly when handling payloads. We qualify this class of behaviors, where whole-body locomotion and manipulation are jointly executed to accomplish forceful tasks, as Dynamic Mobile Manipulation (DMM). Despite rapid advancements in perception and controls, robots still struggle to execute these maneuvers reliably.

Recently, learned policies have demonstrated strong potential in enabling mobile manipulation tasks, such as lifting boxes. 
Authors of \cite{BoxLocoManSimToReal} employ diffusion policies and sim-to-real transfer to successfully demonstrate humanoid box loco-manipulation on hardware. Others have also explored training policies using human demonstrations collected through teleoperation, enabling humanoid robots to perform loco-manipulation tasks based on imitation learning \cite{ImitationLearningHumanoid}. As the need for collecting human demonstration data for learning grows, the ability to teleoperate robots for DMM tasks becomes more pertinent. Teleoperation presents a pathway to teaching robots DMM-like tasks by embedding the human pilots into the robot control loop.

Motion retargeting strategies allow the robot to behave as an extension of the human, where the human pilot uses their arms and body to simultaneously control the robot's manipulation and locomotion for adaptive tasks. Haptic feedback may also be applied to the operator to enhance the teleoperation experience by allowing the pilot to feel the robot's interaction forces, improving the quality and authenticity of the demonstrations \cite{ImitiationLearningHaptics}.

\begin{figure}
    \centering
    \includegraphics[width = \columnwidth]{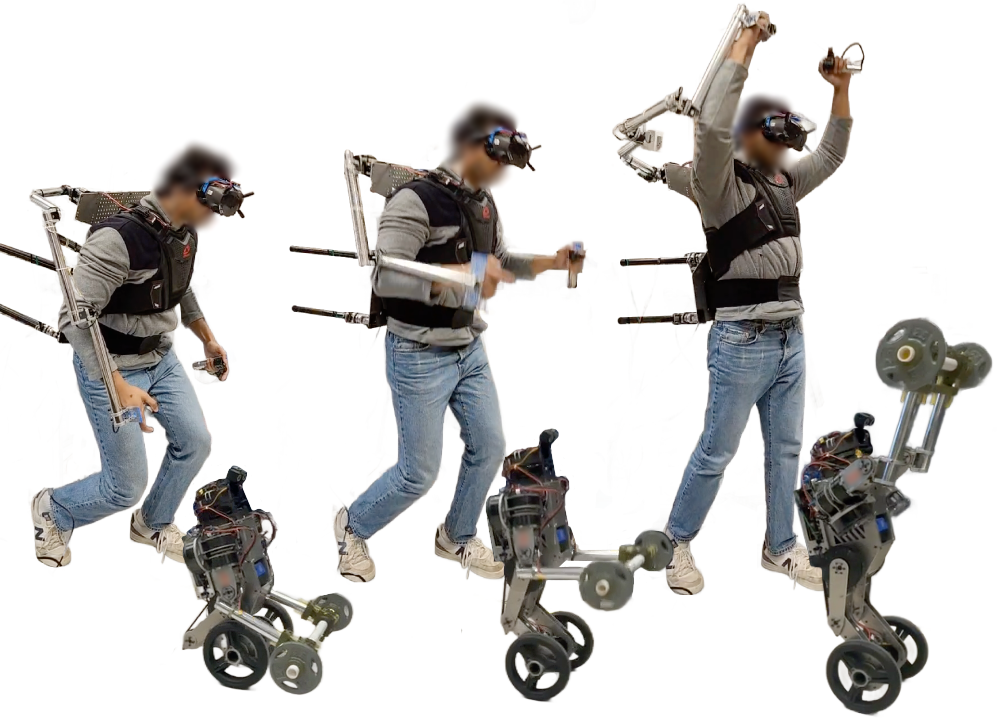}
    \caption{A human operator controls the robot to use its whole body to lift a barbell 21\% of its weight. The robot automatically leans backward to compensate for the additional weight, allowing the human to remain mostly upright.}
    \label{fig:snatch}
\end{figure}

Prior work has demonstrated elements of robot whole-body control for heavy lifting tasks. Bipedal humanoid platforms have used trajectory optimization to generate lifting motions \cite{Humanoid_lifting_task, Humanoid_lifting_task_2, HumanoidLiftingHead} but have only been shown for tasks where the robot is standing in one place. Wheeled humanoids such as Golem Krang \cite{stilman2010golem} have shown promise in tasks requiring locomotion, and height change but did not show any manipulation. The authors in \cite{CMU_ballbot} show Ballbot adjusting for the added payload on its single DoF arm by adjusting its lean angle, but remains at a fixed height. 

 Teleoperated platforms, such as CENTAURO \cite{TeleopSharedHeavyLift}, have explored lifting, but focus largely on estimating grasp forces. Their demonstrated payloads are small relative to the robot's body mass, and the systems do not dynamically reconfigure their posture during the task. Recently, the Reflex teleoperated mobile robot showed impressive locomanipulation capabilities while moving, bending down and lifting heavy bags of rice \cite{ReflexRobotics}. However, this platform is fully actuated and does not explicitly account for disturbance moments on the base created by large payloads during manipulation\cite{ReflexRobotics}. To the best of our knowledge, no prior work has demonstrated teleoperated DMM with height retargeting, explicit handling of payload disturbances, and immersive haptic feedback in a unified system.

In this work, we address these gaps by enabling a human pilot to teleoperate a wheeled humanoid robot to pick up and carry heavy objects while dynamically adjusting its height and base posture. We use a modified version of the Human-Machine Interface shown in \cite{wang2021dynamic} to deliver haptic forces and moments, measure the pilot's arm motion, height, pitch, and center of pressure. Retargeting these human states to desired robot setpoints, the pilot can then control the whole body of the robot. To balance itself while lifting heavy objects, the robot must adjust its lean by automatically compensating for the mass at the controller level or by relying on the the human to make the needed adjustment. These control strategies lead to varied teleoperation experiences. As such, we evaluate three different mappings for telelocomotion control—ranging from manual to automated balance compensation—to assess pilot preferences in tasks involving barbell and box lifting.

The primary contributions of this work are:
\begin{itemize}
    \item Human-motion retargeting for heavy-lifting tasks with height-varying control.
    \item A pitch-based locomotion retargeting strategy in which the pilot regulates the linearized error dynamics of the robot during heavy-object lifting.
    \item An experimental comparison of two disturbance rejection paradigms: pilot-compensation versus robot controller compensation, evaluated for heavy box-lifting tasks on hardware.
\end{itemize}

\section{Background}
\label{sec:background}

Use of whole-body human motion for controlling bipedal or wheeled humanoids has shown feasibility across a range of tasks \cite{nakaoka2003generating, wang2021dynamic}. Modeling the human as a fixed-base inverted pendulum, we define pitch, $\theta_H$, as the angle between the human ankle and their center of mass (CoM), as shown in Fig. \ref{fig:robot_rom_arms}. The robot is modeled as a wheeled inverted pendulum, and its linearized dynamics along the $x$-axis are given by:
\begin{align}
    \bm{\dot{q}}_{R} = \bm{Aq}_{R} + \bm{Bu}_{R} + \bm{d}_R
\end{align}
where $q_R = [x_R, \theta_R, \dot{x}_R, \dot{\theta}_R]$ is the state vector comprising the wheel position, $x_R$, the robot's pitch angle $\theta_R$ (defined between the wheel axle and the center of mass), and their respective velocities. The term, $\bm{d}_R$ captures external disturbances acting on the robot, including forces and moments.

Here, we review two locomotion retargeting strategies using either base velocity control or a direct pitch control of the robot. We briefly discuss the balance implications of each strategy and how they support heavy lifting tasks. A velocity-based mapping strategy \cite{PuruHumanoids} enables precise base position control by mapping human pitch to desired robot velocity, $\dot{x}_R^{des} \propto \theta_H$. This mapping is well-suited for slower, deliberate positioning but lacks responsiveness as the reference for the base is not consistent with the pitch dynamics of the robot wheeled inverted pendulum. In contrast, a pitch-only based balancing strategy \cite{PuruDMM} directly equates the the robot’s Divergent Component of Motion (DCM), $\xi_R = \theta_R + \dot{\theta}_R/\omega_R$, to the human’s DCM, $\xi_H = \theta_H + \dot{\theta}_H/\omega_H$, where the natural frequencies of the human and robot are given by $\omega_H = \sqrt{g/h_H}$ and $\omega_R = \sqrt{g/h_R}$, respectively. This mapping aligns the robot’s pitching behavior with that of the human, directly retargeting their lean and angular rate. This allows the pilot to dynamically balance the robot in place by modulating their own pitch and rate of change. However, this comes at the cost of reduced position control fidelity and increased cognitive load, as the pilot must manage the motion of the base indirectly through the pitching of the robot. For the remainder of this paper, we refer to this motion retargeting strategy as DCM dynamic similarity (between robot and human pendular models).

These locomotion strategies also influence how balance is maintained during lifting tasks. To safely lift heavy objects positioned in front of the robot, its CoM must stay above the wheelbase. This is accomplished by commanding a backward lean, ensuring that the combined CoM of the robot and object remains aligned with the center of the wheels. To achieve this lean, the pilot may lean back manually or rely on the controller to compensate and automatically adjust its desired pitch \cite{CMU_ballbot}. 

Utilizing the the DCM motion retargeting strategy enables us to explicitly model external forces on the robot along its sagittal plane of motion as a haptic feedback to the pilot that minimizes motion mismatch and conveys a scaled external forces measured at the robot’s end effector:
\begin{equation} \label{eq:dmm_force_fb}
F_{xH}^{HMI} = \gamma_H (\xi_R - \xi_H) + \frac{\gamma_H}{\gamma_R}F_{xR}^{ext}
\end{equation}
where $\gamma_H$ and $\gamma_R$ are non-dimensional gains that scale DCM error and external forces based on human-robot size differences. This feedback supports coordinated whole-body control, aiding the pilot in tasks like object manipulation and push recovery. 
\section{Method} \label{sec:methods}
In prior work \cite{PuruDMM}, DMM tasks were only performed at fixed heights, and only contact forces along the $x$-axis (sagittal plane motion) and moments about the robot’s $z$-axis (yaw) were explicitly modeled. However, moments about the wheel's $y$-axis induced by carrying a payload were not considered in the robot’s control strategy. Rather, the pilot was expected to compensate for the external moment by adjusting their lean backwards. While this approach was feasible for light weights (under 1.25 kg), the pilot did not have enough workspace range of motion to lean far back enough to compensate for heavier weights. Moreover, this extended motion imposed a cognitive load on the pilot for executing tasks that required more precise loco-manipulation. To address these issues, we outline our motion retargeting strategy for adjusting height of the robot, as well as automatically adjusting its desired pitch to compensate for the payload. We compare and contrast applying this disturbance moment as a part our haptic teleoperation framework versus direct compensation in our locomotion controller with minimal feedback to the pilot, as shown in Fig. \ref{fig:haptics_layout_options}. 

Section \ref{method:HeightRetarget}, outlines our locomotion retargeting for changing the robot's height, with an improved locomotion controller. Section \ref{method:new_des_pitch} shows our process for estimating the robot's new pitch setpoint angle based on the known mass of the object. Finally, Section \ref{method:dyn_loco_retarget} outlines an approach to enable the human to control the error between the robot and its new setpoint by equating the dynamics of the human and robot pendulums, resulting in a new feedforward and feedback signal.

\subsection{Height Variation Locomotion Retargeting} \label{method:HeightRetarget}
To enable height variation we allow both human and robot pendular models to have variable pole length. However, the dynamics along the vertical ($z$ axis) are not explicitly modeled. To enable intuitive height control, we map changes in human height to changes in robot height via a kinematic relationship. Specifically, the deviation from nominal human height, $\Delta h_H := h_H - h_H^{nom}$, is mapped to a corresponding change in robot height, $\Delta h_R := h_R - h_R^{nom}$:
\begin{equation}
    \frac{\Delta h_R}{h_R^{nom}} = \frac{\Delta h_H}{h_H^{nom}}
\end{equation}
where $h_R^{nom}$ and $h_H^{nom}$ are the starting height of the human and robot at their upright positions. Rearranging, we can define the desired robot height to be:
\begin{equation} \label{eq:height_mapping}
    h_R^{des} = h_R^{nom} + \beta_{z}\frac{h_R^{nom}}{h_H^{nom}}\Delta h_H 
\end{equation}
where $\beta_z \in [0,1]$ represents a tuning gain that affects how much the human has to bend to change the robot's height. If $\beta_z = 1$, the human's height change as a percentage is the same that of the robot's. For our experiments $\beta_z = 0.5$.
\begin{figure}
    \centering
    \includegraphics[width=1.0\linewidth]{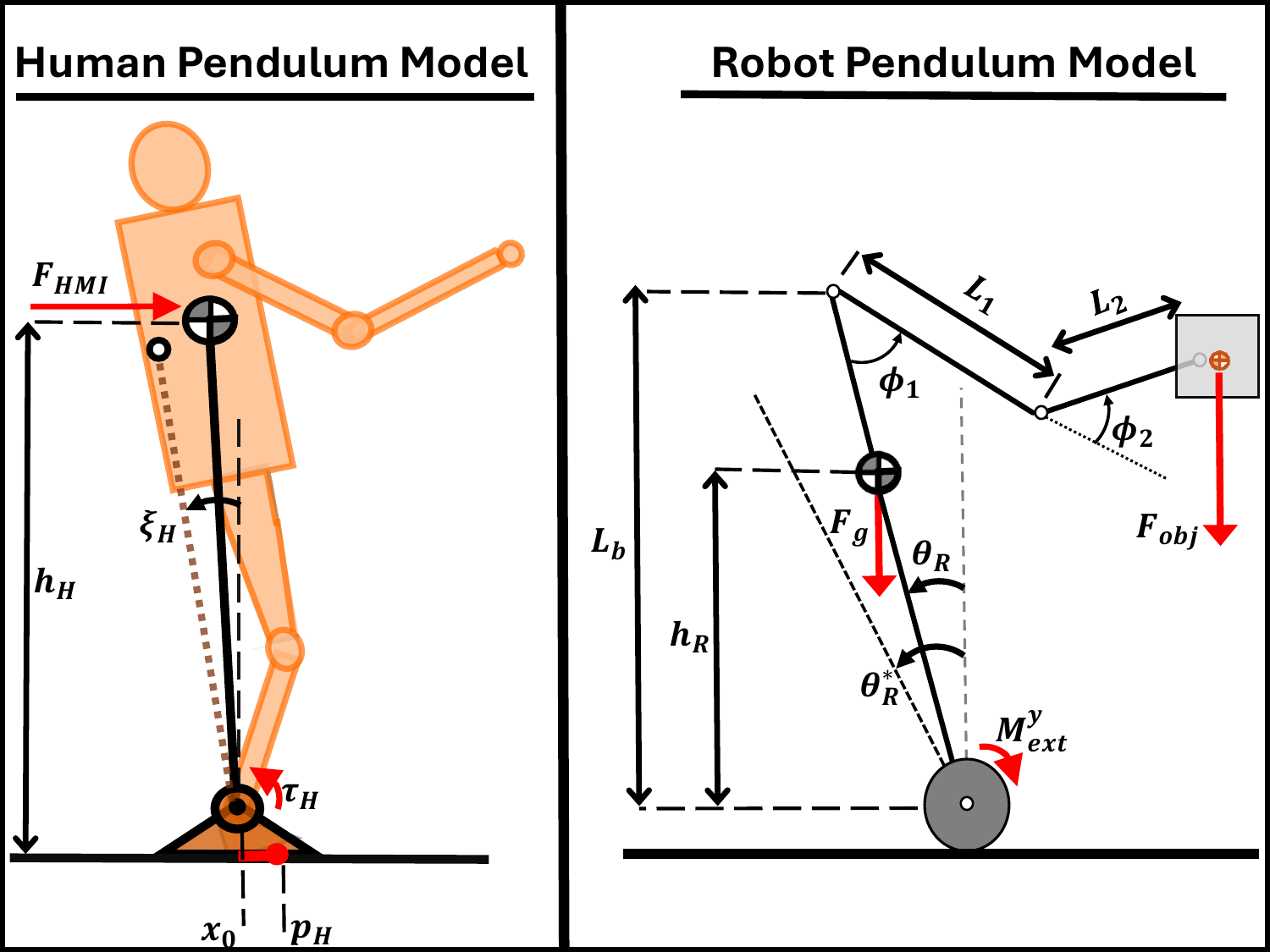}
    \caption{\textit{Left}: The haptic force feedback is generated to produce a moment around the ankle of the human pendular model. \textit{Right}: The robot reduced model computes a new desired pitch angle to counteract the external moment from the object mass.}
    \label{fig:robot_rom_arms}
\end{figure}

When changing height and applying a ground reaction force, the robot base may experience a component of the applied leg force in $x$ direction due to oscillations of the robot when balancing. This undesirable disturbance from the legs on the robot base is given by the current robot pitch and estimated ground reaction force between the contact point and the CoM, $F_L$:
\begin{equation}
    \bm{d}_w = \begin{bmatrix}
        F_Lsin(\theta_R) \\
        0
    \end{bmatrix} 
\end{equation}
Consequently, we design our feedback controller to include  feedback linearization to explicitly compensate for the effects of these change in height dynamics, and an gain-scheduled LQR \cite{wheeledhumanoid2019model} to balance the robot at various heights:
\begin{align} \label{eq:LQR_control}
    u = -\bm{K}_{LQR}(h_R)(\bm{q}_{xR}^{des} - \bm{q}_{xR}) - \bm{B}^{\dagger}\bm{d}_w
\end{align}
where the LQR gain, $\bm{K}_{LQR}$, is linearly interpolated between gain matrices as a function of the robot height, and $\bm{B}^{\dagger}$ is the pseudo-inverse of the input matrix.

\subsection{Robot Desired Pitch Estimation with Payload} \label{method:new_des_pitch}
When lifting heavy objects, an external moment is induced around the base of the robot as a function of the robot's pitch, and distance of the object from the wheel's center. Shifting the CoM beyond the support line between the robot's wheels causes the robot to begin falling. To account for the shifted mass, we solve for the desired lean of the robot in static equilibrium, $\theta_R^*$, while modeling the robot arm as a planar manipulator fixed at the robot's shoulder as shown in Fig. \ref{fig:robot_rom_arms}. The sum of moments ($M_g ^y+ M_{ext}^y = 0$) around the wheel are given by:
\begin{align} \label{eq:sum_of_moments}
    F_g h_R s_{\theta_R^*} &= -F_{obj}x_{ee}
\end{align}
where $F_g$ and $F_{obj}$ are the forces due to gravity on the robot and object, respectively, and $s_{\theta^{*}_R}$ denotes $sin(\theta_R^*)$. The robot's end-effector position along the x axis, $x_{ee}$, is given by:
\begin{equation} \label{eq:rom_xee}
    \begin{aligned}
        x_{ee} = L_b s_{\theta_R^*} + L_{1}s_{\theta_R^* + \phi_1} + L_{2}s_{\theta_R^* + \phi_1 + \phi_2}
    \end{aligned}
\end{equation}
where $\phi_1$, $\phi_2$ are the arm joint angles, $L_b$ the length from the robot base to shoulder, $L_1$ the length of the upper arm link, and $L_2$ the length of the forearm link. The arms are modeled as planar here, but can easily also be expanded for the 3D case to use the full forward kinematics of the end effector.

Substituting Eq. \ref{eq:rom_xee} into Eq. \ref{eq:sum_of_moments} we can solve for the desired pitch angle for the robot:

\begin{equation} \label{eq:equil_pitch}
    \begin{aligned}
        \theta_R^* &= \tan^{-1}({\frac{f_1}
        {f_2}}) \\
        f_1 &= -F_{obj}(L_1 s_{\phi_1} + L_2(s_{\phi_1}c_{\phi_2} + c_{\phi_1}s_{\phi_2}) \\
        f_2 &= F_g h_R \!+\! F_{obj}(L_b \!+\! L_1 c_{\phi_1} \!+\! L_2(c_{\phi_1}c_{\phi_2} \!-\! s_{\phi_1}s_{\phi_2})) 
    \end{aligned}
\end{equation}
To improve tracking, we solve for the desired pitch velocity by taking the time derivative of Eq. \ref{eq:equil_pitch}:
\begin{align} \label{eq:equil_pitch_vel}
    \dot{\theta}^{*}_R = \frac{f_2\frac{df_1}{dt} - f_1\frac{df_2}{dt}}{f_1^2 + f_2^2}
\end{align}
where $f_1(\phi_1, \phi_2, F_{obj})$ and $f_2(\phi_1, \phi_2, F_{obj})$ are differentiable functions in Eq.\ref{eq:equil_pitch}. This desired pitch velocity changes as a function of how quickly the robot moves its arm so is dependent on the arm joint velocities. The desired pitch and pitch velocities are then tracked by the controller in Eq. \ref{eq:LQR_control}.

\subsection{Locomotion Retargeting of Error Dynamics} \label{method:dyn_loco_retarget}

To give the pilot pitch based locomotion control of the robot, we build upon the methodology in \cite{PuruDMM}, where the tracking of the human-robot DCM was utilized $\xi_R = \xi_H$. In these works, however, the robot was linearized around the upright $\theta_R^* = 0$ position and the desired setpoint for the controller was solely modulated by the human reference. 

When lifting the box, the system dynamics switch discretely to one including the robot dynamics and the object dynamics. To account for the shift in setpoint, we allow the robot to automatically adjust its lean by regulating toward a desired DCM, computed using the desired pitch angle and pitch velocity defined in Eqs. \ref{eq:equil_pitch} and \ref{eq:equil_pitch_vel}.
\begin{equation}
    \xi_{R}^* = \theta_{R}^* + \frac{\dot{\theta}_{R}^*}{\omega_R}
\end{equation}
To give the pilot control of the robot around the desired DCM, and minimize tracking error we enforce that the frequency normalized DCM velocity of the robot around its new setpoint and that of the human be the same:
\begin{equation}
    \frac{\dot{\xi}_R - \dot{\xi}_R^*}{\omega_R} = \frac{\dot{\xi}_H}{\omega_H}
\end{equation}
In other words, the error between the dynamics of the robot and a reference robot model, $\dot{\xi}_R^*$, that accounts for the payload are controlled by the human. Expanding the equations similar to \cite{PuruDMM}, we find a new feedback, $F_{fb}$ and feedfoward, $F_R$:
\begin{equation} \label{eq:dyn_sim}
    \begin{aligned}
        F_{fb} \!=\! \gamma_H \big((\theta_R &\!-\! \theta_R^* \!-\! \theta_H) \!+\! (\frac{\dot{\theta}_R \!-\! \dot{\theta}_R^*}{\omega_R} \!-\! \frac{\dot{\theta}_H}{\omega_H})\big) \\
        F_{ff} &= \gamma_R (F_{ff}^* + \frac{p_H}{h_H})
    \end{aligned}
\end{equation}
where $\gamma_H$ and $\gamma_R$ are non-dimensionalizing coefficients that account for the human and robot mass and height. In this framework, the natural frequency and height of the robot are updated based on the retargeted height from the previous section. We assume that the human's change in height is much slower than the controller update frequency. As such, we treat the human height as constant in any single iteration of the controller's linearization, and fix the natural frequency. 

In practice, we treated the reference robot model as passive by setting the desired cart force to $F_{ff}^* = 0$. Pilots preferred this approach, as it allowed for more responsive dynamic motions controlled through the human's pitch.
\begin{figure}
    \vspace{0.75em}
    \centering
    
    \includegraphics[width=1.0\linewidth]{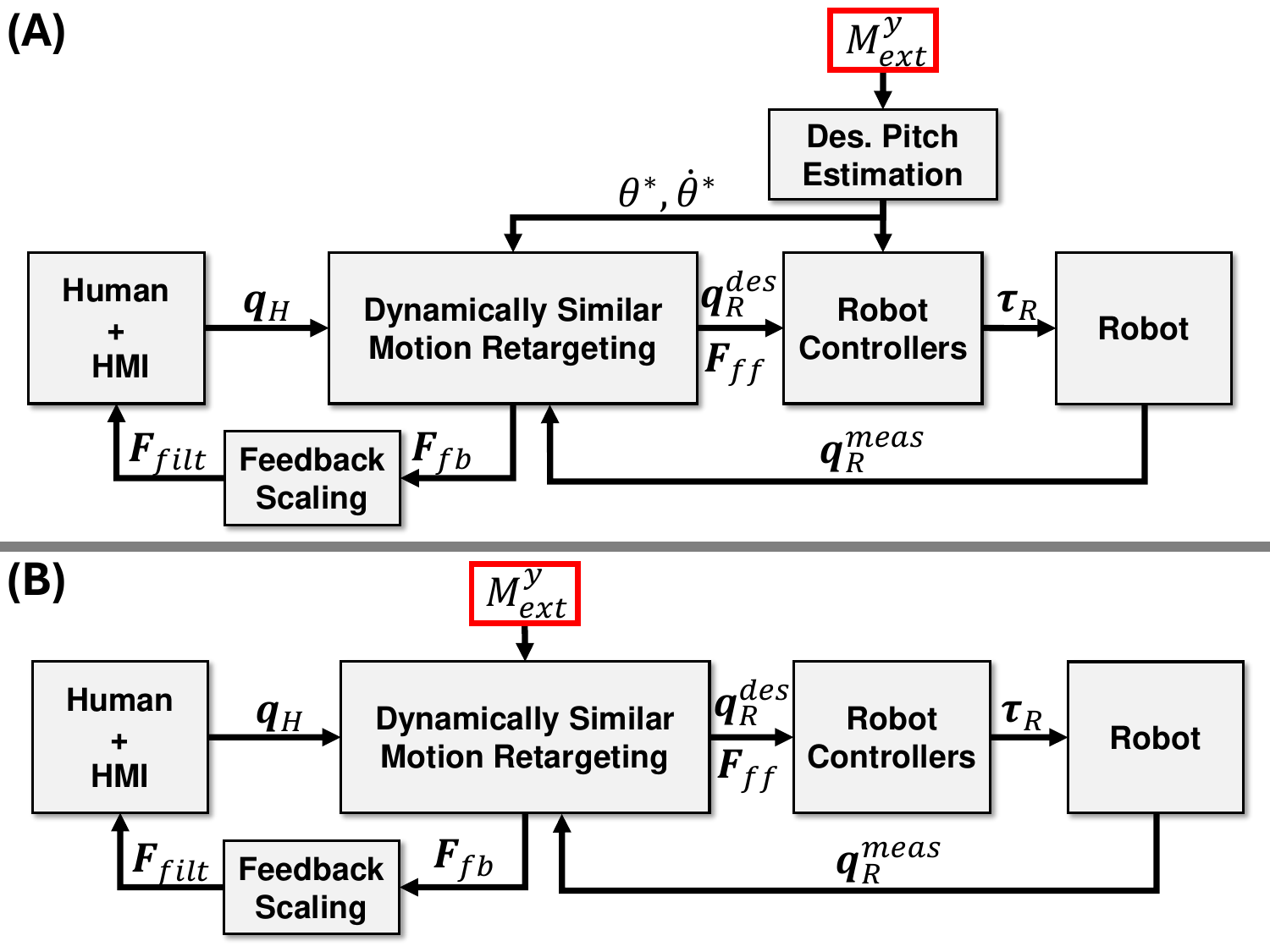}
    \vspace*{-2em}
    \caption{(A) highlights control layout for experiments where the robot automatically compensates for the external moment by estimating a new desired pitch. (B) shows the layout where the moment is fed back entirely to the pilot through dynamic similarity retargeting.}
    \label{fig:haptics_layout_options}
    \vspace{-1.25em}
\end{figure}

\begin{figure*}[t] 
    \vspace{0.75em}
    \centerline{\includegraphics[width=17.5cm]{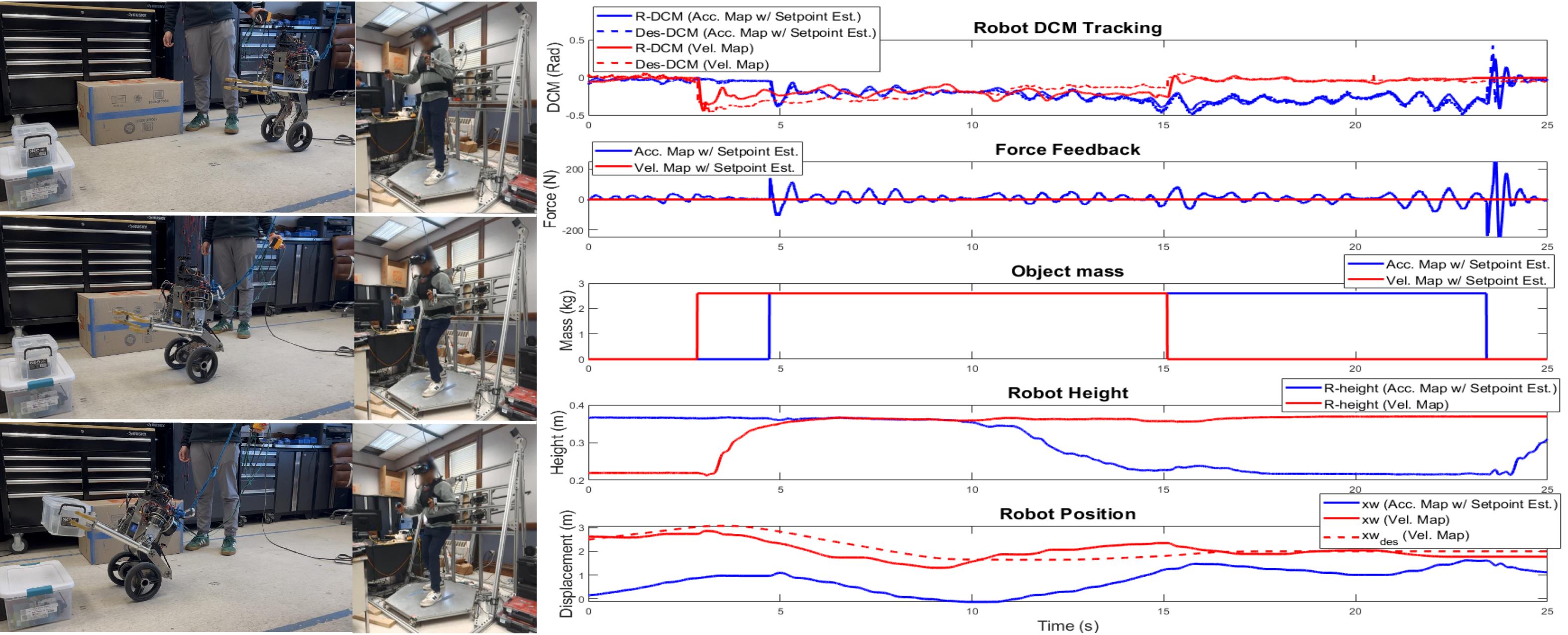}}
    \vspace*{-0.5em}
    \caption{\textit{Left}: A pilot operates the robot to pick and place a 2.5 kg (21 \% of $m_R$) box at different heights. \textit{Right}: Plots comparing the DCM pitch-based mapping (blue) with the baseline velocity mapping (red). The DCM mapping with force feedback results in better DCM tracking. Note that in this experiment, the robot grasps the box at an elevated height and places it at a lower height using the DCM pitch mapping, while the opposite is performed using the velocity mapping. Similar results are obtained for the reverse order.}
    \label{fig:BoxLiftData}
    \vspace{-1.25em}
\end{figure*}

\section{Experimental Setup} \label{sec:exp_setup}

We used three different control strategies over 2 sets of experiments to validate robot execution of the task and pilot preferences for transferring a heavy weight (21\% of the robot’s mass) between elevated and lowered positions. In all experiments we assumed that the object mass (accurate within 10\% of the actual value) is known apriori. The developed whole-body teleoperation framework is now enhanced with the ability to change the robot’s height, reflect the induced moment felt around the robot's base to the pilot, and automatically compensate for this moment through tracking of a new pitch angle setpoint.

Our baseline control strategy relied on base-velocity retargeting, mapping the human’s pitch to the robot’s base velocity ($\theta_H \rightarrow \dot{x}_R$) \cite{PuruHumanoids} and used a robot controller that compensated for the payload by adjusting its pitch setpoint. This setup served as a benchmark for evaluating the impact of haptic feedback and pilot preferences in the absence of direct pitch control.

Alternatively, the pilot could use the proposed DCM pitch-only mapping, which included automatic compensation for the robot’s pitch due to the payload. The pilot activates this mode via a right hand trigger on the HMI and activates pitch compensation during the lift using a left hand trigger. Once the new pitch setpoint was defined, the locomotion controller updated its reference to support balance. Haptic feedback in this mode reflected the error between the human's DCM and robot's DCM relative to its new setpoint, as defined in Eq.\ref{eq:dyn_sim} and illustrated in Fig.\ref{fig:haptics_layout_options}(A).

Finally, we tested a fully manual control strategy, derived from the DCM pitch-only mapping but with no automatic pitch compensation. In this mode, the haptic force conveyed the moment induced around the robot’s base, scaled by the human and robot parameters:
\begin{equation} 
F_{fb} = \frac{\gamma_H}{\gamma_R}M_{ext}^y 
\end{equation} 
where $M_{ext}^y$ is estimated from the right side of Eq.\ref{eq:sum_of_moments}. This mapping directly reflected the full payload-induced moment to the pilot, requiring them to manually adjust the robot’s lean to maintain balance. The moment caused by the payload was treated as an external disturbance within the motion retargeting framework, as illustrated in Fig.\ref{fig:haptics_layout_options}(B).

\section{Results \& Discussion} \label{sec:results}

Section \ref{Exp:setpoint_est_prefernce} details experiments in which pilots were asked to lift a heavy object and relocate it to a different height using two control modalities: one providing full haptic feedback from the robot (manual mode) and the other employing an automatic setpoint estimation for lean compensation. Section \ref{Exp:Mapping_prefernce} then examines the efficacy and user preference between controlling the velocity of the robot base versus directly controlling the robot pitch motion and DCM during heavy box lifting. All experiments were conducted in compliance with the university Internal Review Board (IRB) protocols and are illustrated in accompanying video footage.

\subsection{Setpoint Estimation Preference} \label{Exp:setpoint_est_prefernce}

In this experiment, the primary task was to lift a 2.5 kg object and place it at a target height. Our framework allowed the pilot to simultaneously control the robot’s forward motion and adjust its height, eliminating the need for switching between distinct locomotion and stance modes. 

\begin{figure}[t]
    \vspace{0.75em}
    \centering
    \includegraphics[width=1.0\linewidth]{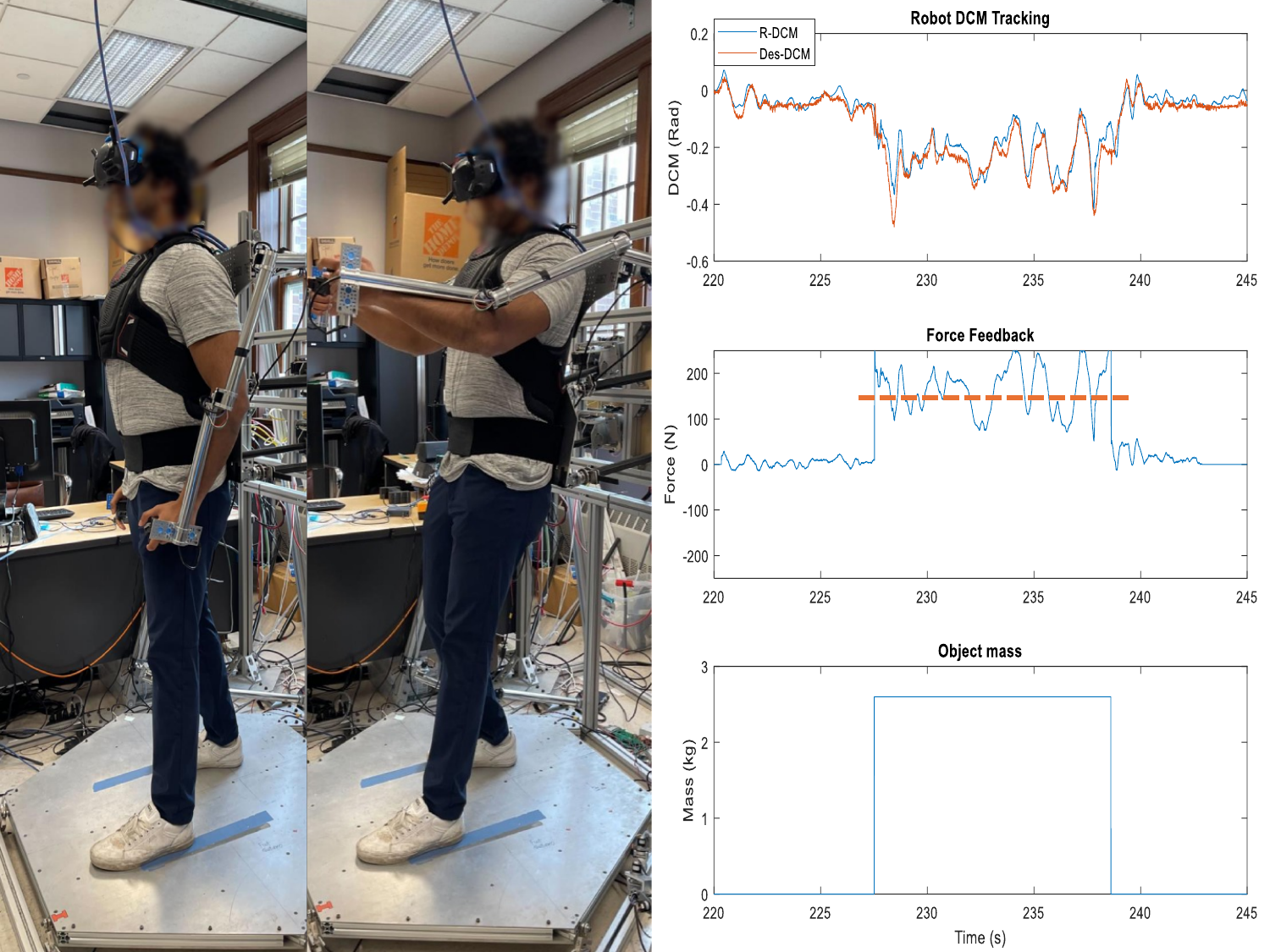}
    \vspace*{-2em}
    \caption{Without setpoint estimation, the pilot must resist nearly 200 N of haptic force to balance the robot holding a heavy object.}
    \label{fig:manual_control}
    \vspace{-1.25em}
\end{figure}

When operating in manual mode, where the full haptic moment feedback is provided, pilots naturally adjusted their grasp by drawing the box closer to the robot’s body. This strategy reduced the net moment around the base and decreased the force the user needed to apply to the ground to counteract the external moment. DCM tracking performance decreased during lifting, and the pilot was forced to lean back excessively via the HMI to compensate for the weight as shown in Fig. \ref{fig:manual_control}. This scenario underscores a critical limitation of manual lean control: the pilot’s available range of motion is insufficient to counterbalance heavier loads.

In contrast, the automatic setpoint estimation mode was consistently favored by the pilots. With this approach, the system autonomously compensated for the robot’s leaning, thereby reducing the feedback amplitude that the pilot needed to counteract. In the manual mode, the necessity to exert large corrective forces over prolonged periods rendered the operation unsustainable, as the pilots were forced to constantly lean to balance the external moments.

These experiments also demonstrated the benefit of introducing a tunable gain on the haptic feedback: $F_{fb}~\leftarrow~K_{fb}F_{fb}$. Setting $K_{fb} = 0$ results in open-loop control with no feedback, while $K_{fb} = 1$  provides full, high-fidelity force feedback. While full feedback enhances transparency, it can trigger oscillations if the pilot becomes overwhelmed and their motions are overly governed by the haptic force. An intermediate gain allowed pilots to adjust feedback strength for better stability, comfort, and tracking performance.

\subsection{Mapping Preference for Heavy Box Lift} \label{Exp:Mapping_prefernce}

These experiments compare pilot preferences between base velocity control and DCM pitch-only control when lifting a heavy box. Both control schemes utilized automatic setpoint estimation to determine the desired robot pitch, but differ in responsiveness.

The base velocity mapping provides a more conservative control interface by tracking the robot’s wheel position, while keeping an upright robot pitch. This can lead to decreased tracking of large desired pitch angles from the human. In contrast, the DCM mapping offers a more dynamic response by directly commanding the robot’s acceleration through its pitch. Although both mappings enabled successful task execution, pilots reported distinct subjective preferences. 

Early in the trials, pilots tended to favor the velocity-based interface, likely due to its predictability and the minimal adjustments required for precise grasping and box placement. As familiarity with the task increased, they increasingly transitioned to the DCM mapping to exploit its enhanced responsiveness for dynamic base motions. When releasing the object, pilots frequently reverted to the velocity mapping. This switch neutralized the haptic feedback and helped to minimize unwanted oscillations in the robot’s movements.

The experiments indicate that pilots benefit from a hybrid control strategy where the choice between base velocity or DCM pitch mappings are dynamically selected based on task phase and individual operator preference. This adaptive approach allows a tradeoff between precise positioning during grasping and releasing, increasing responsiveness in locomotion, and minimizing cognitive load in idle states.

\subsection{Limitations}

One limitation of our method is that it requires an estimate of the object's mass prior to grasping it, in order to compute a new robot pitch equilibrium as soon as the grasp is complete. Existing methods can estimate the mass of the object shortly after interacting with the object \cite{DonghoonParamID}, but it remains a challenge to compensate for object mass during the grasp. Also, we model the object as a point mass held at the end effector; objects with a center of mass offset in the x-direction may cause inaccurate compensation, and objects with a center of mass offset in the y-direction may cause the robot to roll. Future work may use vision-based estimation of the object's inertial parameters before grasping.

Another limitation of our current system is grasping inconsistency. The robot uses passive grippers that apply opposing forces to the sides of the object, relying on friction and small protrusions to hold the object. Some object geometries require precise alignment, and we find that the lack of depth perception with the monocular first-person-view (FPV) cameras often results in an error in perceiving whether the robot grippers are aligned with the object.   

\section{Conclusion} \label{sec:conclusion}
We develop a teleoperation framework that enables dynamic mobile manipulation on a wheeled humanoid robot, integrating height retargeting, explicit handling of payload mass, and haptic feedback. The system allows a human pilot to control the robot’s whole-body posture and locomotion using motion retargeting, while receiving force cues that reflect external interaction forces and balance demands without overwhelming the user. Through heavy lifting experiments, we showed that automatic lean compensation improved task performance and reduced pilot effort compared to manual feedback. We also found that base velocity control and pitch-based control provide complementary benefits across different phases of the task, depending on precision requirements.


\appendices
\bibliographystyle{IEEEtran}
\bibliography{Main.bib}

\begin{thebibliography}{10}
\providecommand{\url}[1]{#1}
\csname url@samestyle\endcsname
\providecommand{\newblock}{\relax}
\providecommand{\bibinfo}[2]{#2}
\providecommand{\BIBentrySTDinterwordspacing}{\spaceskip=0pt\relax}
\providecommand{\BIBentryALTinterwordstretchfactor}{4}
\providecommand{\BIBentryALTinterwordspacing}{\spaceskip=\fontdimen2\font plus
\BIBentryALTinterwordstretchfactor\fontdimen3\font minus \fontdimen4\font\relax}
\providecommand{\BIBforeignlanguage}[2]{{%
\expandafter\ifx\csname l@#1\endcsname\relax
\typeout{** WARNING: IEEEtran.bst: No hyphenation pattern has been}%
\typeout{** loaded for the language `#1'. Using the pattern for}%
\typeout{** the default language instead.}%
\else
\language=\csname l@#1\endcsname
\fi
#2}}
\providecommand{\BIBdecl}{\relax}
\BIBdecl

\bibitem{BoxLocoManSimToReal}
J.~Dao, H.~Duan, and A.~Fern, ``Sim-to-real learning for humanoid box loco-manipulation,'' in \emph{2024 IEEE International Conference on Robotics and Automation (ICRA)}, 2024, pp. 16\,930--16\,936.

\bibitem{ImitationLearningHumanoid}
M.~Seo, S.~Han, K.~Sim, S.~H. Bang, C.~Gonzalez, L.~Sentis, and Y.~Zhu, ``Deep imitation learning for humanoid loco-manipulation through human teleoperation,'' in \emph{2023 IEEE-RAS 22nd International Conference on Humanoid Robots (Humanoids)}.\hskip 1em plus 0.5em minus 0.4em\relax IEEE, 2023, pp. 1--8.

\bibitem{ImitiationLearningHaptics}
C.~Cuan, A.~Okamura, and M.~Khansari, ``Leveraging haptic feedback to improve data quality and quantity for deep imitation learning models,'' \emph{IEEE Transactions on Haptics}, 2024.

\bibitem{Humanoid_lifting_task}
H.~Arisumi, S.~Miossec, J.-R. Chardonnet, and K.~Yokoi, ``Dynamic lifting by whole body motion of humanoid robots,'' in \emph{2008 IEEE/RSJ International Conference on Intelligent Robots and Systems}, 2008, pp. 668--675.

\bibitem{Humanoid_lifting_task_2}
K.~Harada, S.~Kajita, H.~Saito, M.~Morisawa, F.~Kanehiro, K.~Fujiwara, K.~Kaneko, and H.~Hirukawa, ``A humanoid robot carrying a heavy object,'' in \emph{Proceedings of the 2005 IEEE International Conference on Robotics and Automation}, 2005, pp. 1712--1717.

\bibitem{HumanoidLiftingHead}
R.~Shigematsu, S.~Komatsu, Y.~Kakiuchi, K.~Okada, and M.~Inaba, ``Lifting and carrying an object of unknown mass properties and friction on the head by a humanoid robot,'' in \emph{2018 IEEE-RAS 18th International Conference on Humanoid Robots (Humanoids)}, 2018, pp. 1--9.

\bibitem{stilman2010golem}
M.~Stilman, J.~Olson, and W.~Gloss, ``Golem krang: Dynamically stable humanoid robot for mobile manipulation,'' in \emph{2010 IEEE International Conference on Robotics and Automation}.\hskip 1em plus 0.5em minus 0.4em\relax IEEE, 2010, pp. 3304--3309.

\bibitem{CMU_ballbot}
F.~Sonnleitner, R.~Shu, and R.~L. Hollis, ``The mechanics and control of leaning to lift heavy objects with a dynamically stable mobile robot,'' in \emph{2019 International Conference on Robotics and Automation (ICRA)}, 2019, pp. 9264--9270.

\bibitem{TeleopSharedHeavyLift}
D.~Torielli, L.~Muratore, A.~De~Luca, and N.~Tsagarakis, ``A shared telemanipulation interface to facilitate bimanual grasping and transportation of objects of unknown mass,'' in \emph{2022 IEEE-RAS 21st International Conference on Humanoid Robots (Humanoids)}, 2022, pp. 738--745.

\bibitem{ReflexRobotics}
R.~Robotics, ``Reflex robotics’ wheeled human gives you human-level service,'' \url{https://www.youtube.com/watch?v=Z0komrwHgMA}, 2025, accessed: 2025-05-03.

\bibitem{wang2021dynamic}
S.~Wang and J.~Ramos, ``Dynamic locomotion teleoperation of a reduced model of a wheeled humanoid robot using a whole-body human-machine interface,'' \emph{IEEE Robotics and Automation Letters}, vol.~7, no.~2, pp. 1872--1879, 2021.

\bibitem{nakaoka2003generating}
S.~Nakaoka, A.~Nakazawa, K.~Yokoi, H.~Hirukawa, and K.~Ikeuchi, ``Generating whole body motions for a biped humanoid robot from captured human dances,'' in \emph{2003 IEEE International Conference on Robotics and Automation (Cat. No. 03CH37422)}, vol.~3.\hskip 1em plus 0.5em minus 0.4em\relax IEEE, 2003, pp. 3905--3910.

\bibitem{PuruHumanoids}
A.~Purushottam, J.~Yan, C.~Xu, Y.~Sim, and J.~Ramos, ``Wheeled humanoid bilateral teleoperation with position-force control modes for dynamic loco-manipulation,'' in \emph{2024 IEEE-RAS 23rd International Conference on Humanoid Robots (Humanoids)}, 2024, pp. 764--771.

\bibitem{PuruDMM}
A.~Purushottam, C.~Xu, Y.~Jung, and J.~Ramos, ``Dynamic mobile manipulation via whole-body bilateral teleoperation of a wheeled humanoid,'' \emph{IEEE Robotics and Automation Letters}, vol.~9, no.~2, pp. 1214--1221, 2024.

\bibitem{wheeledhumanoid2019model}
H.~Zhou, X.~Li, H.~Feng, J.~Li, S.~Zhang, and Y.~Fu, ``Model decoupling and control of the wheeled humanoid robot moving in sagittal plane,'' in \emph{2019 IEEE-RAS 19th International Conference on Humanoid Robots (Humanoids)}.\hskip 1em plus 0.5em minus 0.4em\relax IEEE, 2019, pp. 1--6.

\bibitem{DonghoonParamID}
D.~Baek, B.~Peng, S.~Gupta, and J.~Ramos, ``Online learning-based inertial parameter identification of unknown object for model-based control of wheeled humanoids,'' \emph{IEEE Robotics and Automation Letters}, vol.~9, no.~12, pp. 11\,154--11\,161, 2024.

\end{thebibliography}
\end{document}